\documentclass{article}

\usepackage{microtype}
\usepackage{graphicx}
\usepackage{subcaption}
\usepackage{booktabs} 
\usepackage{threeparttable}

\usepackage{natbib}
\usepackage[utf8]{inputenc} 
\usepackage{hyperref}       
\usepackage{url}            
\usepackage{booktabs}       
\usepackage{amsfonts}       
\usepackage{nicefrac}       
\usepackage{xcolor}
\usepackage{layouts}

\usepackage{amsmath,amsfonts}
\usepackage{amssymb,amsopn}
\usepackage{bm} 
\usepackage{cleveref}
\usepackage{enumerate}

\newcommand{\vct}[1]{\boldsymbol{\mathbf{#1}}} 
\newcommand{\mat}[1]{\boldsymbol{\mathbf{#1}}} 
\newcommand{\sgn}{\text{sgn}} 
\newcommand{\X}{\vct{X}} 
\newcommand{\Z}{\vct{Z}} 
\newcommand{\x}{\vct{x}} 

\newcommand{\z}{\vct{z}} 
\newcommand{\h}{\vct{h}} 
\newcommand{\W}{\mat{W}} 

\usepackage{tikz}
\usetikzlibrary{positioning}
\usetikzlibrary{arrows.meta}

\usepackage{hyperref}

\usepackage[accepted]{icml2021}

\icmltitlerunning{Unsupervised Representation Learning via Neural Activation Coding}

\begin{document}

\twocolumn[
\icmltitle{Unsupervised Representation Learning via Neural Activation Coding}



\icmlsetsymbol{equal}{*}

\begin{icmlauthorlist}
\icmlauthor{Yookoon Park}{columbia}
\icmlauthor{Sangho Lee}{snu}
\icmlauthor{Gunhee Kim}{snu}
\icmlauthor{David M. Blei}{columbia}
\end{icmlauthorlist}

\icmlaffiliation{columbia}{Computer Science Department, Columbia University, New York, USA}
\icmlaffiliation{snu}{Department of Computer Science and Engineering,  Seoul National University, Seoul, South Korea}

\icmlcorrespondingauthor{David M. Blei}{david.blei@columbia.edu}

\icmlkeywords{Machine Learning, ICML}

\vskip 0.3in
]



\printAffiliationsAndNotice{}  

\begin{abstract}
We present \textit{neural activation coding} (NAC) as a novel approach for learning deep representations from unlabeled data for downstream applications.
We argue that the deep encoder should maximize its nonlinear expressivity on the data for downstream predictors to take full advantage of its representation power. 
To this end, NAC maximizes the mutual information between activation patterns of the encoder and the data over a noisy communication channel. We show that learning for a noise-robust activation code increases the number of distinct linear regions of ReLU encoders, hence the maximum nonlinear expressivity. 
More interestingly, NAC learns \textit{both} continuous and discrete representations of data, which we respectively evaluate on two downstream tasks: (i) linear classification on CIFAR-10 and ImageNet-1K and (ii) nearest neighbor retrieval on CIFAR-10 and FLICKR-25K. 
Empirical results show that NAC attains better or comparable performance on both tasks over recent baselines including SimCLR and DistillHash. In addition, NAC pretraining provides significant benefits to the training of deep generative models.
Our code is available at \url{https://github.com/yookoon/nac}. 
\end{abstract}

\section{Introduction}
\begin{figure}[t!]
	\centering
	\begin{subfigure}[t]{0.22\textwidth}
		\centering
		\includegraphics[width=1.0\textwidth]{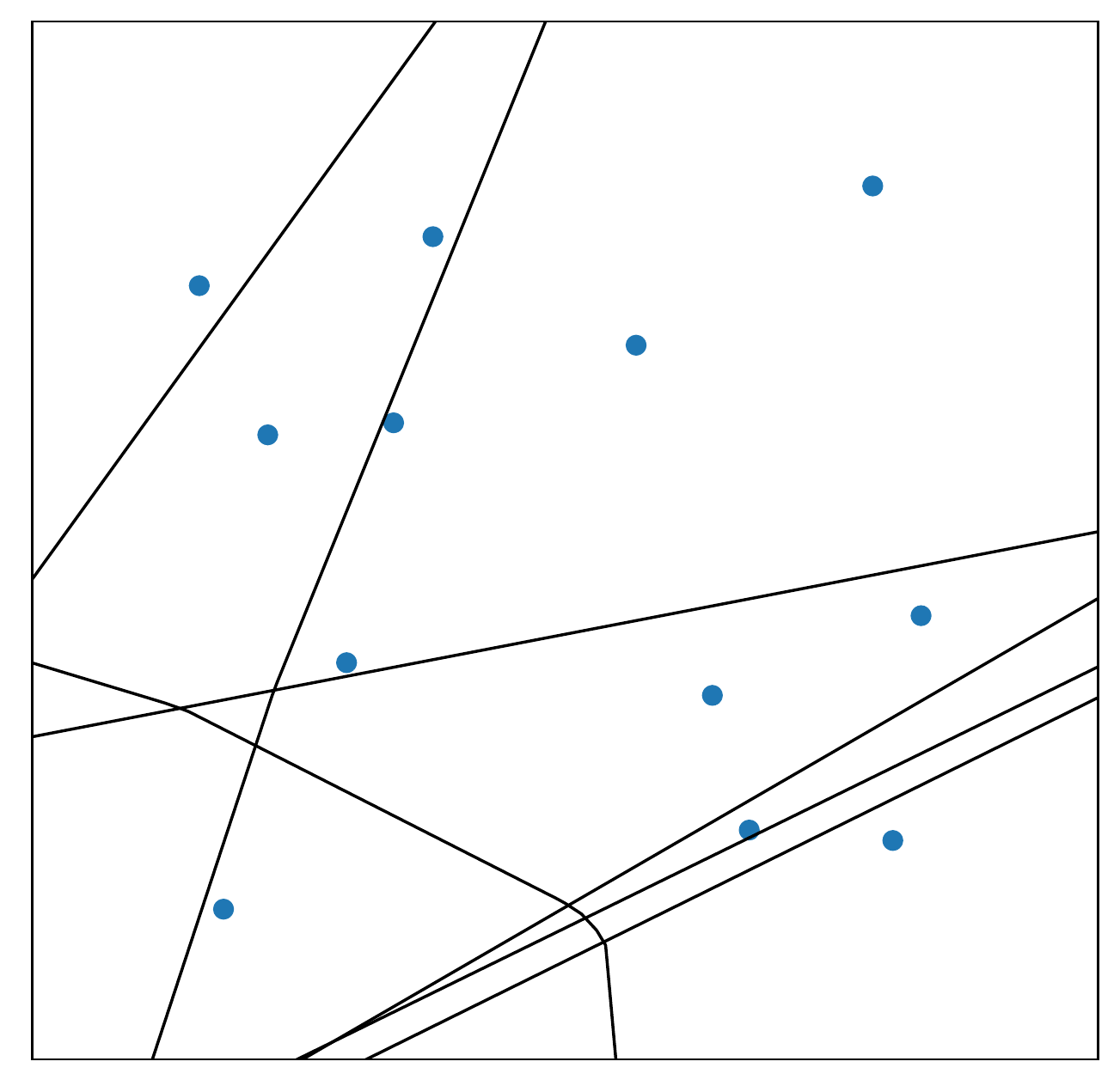}
		\caption{At initialization}
		\label{fig:linear_regions_init}
	\end{subfigure}%
	\hspace{5pt}
	\begin{subfigure}[t]{0.22\textwidth}
		\centering
		\includegraphics[width=1.0\textwidth]{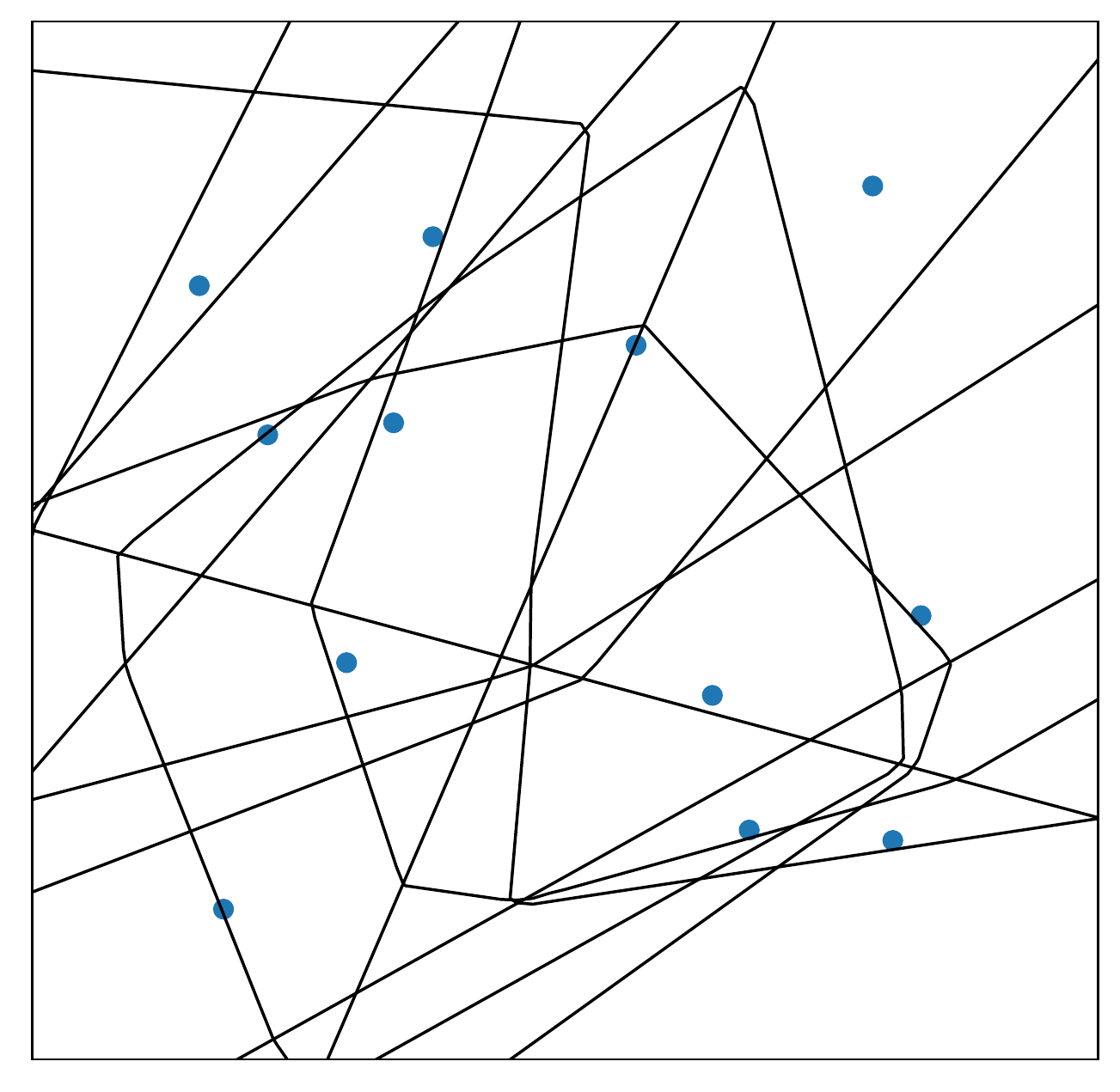}
		\caption{After NAC training}
		\label{fig:linear_regions_trained}
	\end{subfigure}
	\caption{Distinct linear regions of a simple ReLU network with 2 layers of width 4 on 2D toy data. The lines represent the activation boundaries that divide the input space into distinct linear regions. NAC maximizes the number of linear regions of the network on the data, hence the maximum nonlinear expressivity.}
	\label{fig:linear_regions}
\end{figure}	

High dimensional data pose fundamental challenges for machine learning such as the \textit{curse of dimensionality},
and thus often require tailored domain-specific architectures with a large amount of supervision \citep{krizhevsky12,  vaswani2017attention}.
A good representation alleviates such challenges by providing a low-dimensional view of the data that captures high-level semantics and by rendering such information more easily accessible to downstream predictors \citep{bengio2013representation}. 
Especially, unsupervised representation learning possesses a great potential since it provides a means of exploiting abundant unlabeled data for enhancing the performance on downstream applications \citep{devlin2018bert,chen2020simple}, even with limited amounts of labels \citep{chen2020big}. 

We focus on the problem of learning deep representations from unlabeled data for downstream predictors, a popular scenario in unsupervised representation learning literature \citep{wu2018unsupervised, he2020momentum, chen2020simple}.
In this setting, the deep encoder network is pretrained on an unlabeled dataset. The learned representation is then fed into subsequent predictors for downstream tasks such as classification.  
Most often, simple linear models are chosen as the predictors \citep{wu2018unsupervised, he2020momentum, chen2020simple} and the quality of the representation is evaluated by how well these models perform on the downstream applications. 
This evaluation protocol encodes the belief that a good representation should disentangle complex high-level semantics of the data and deliver them in a linearly accessible way. 
The key question here is: how can we learn the deep encoder to benefit the downstream predictors?

Self-supervised learning introduces \textit{pretext} tasks with artificially generated pseudo-labels from unlabeled data \citep{doersch2015unsupervised, noroozi2016unsupervised, gidaris2018unsupervised} to train the encoder, expecting that the encoder would learn useful structures of the data to better solve such tasks. 
Recently, \textit{contrastive learning} of representation \citep{oord2018representation, wu2018unsupervised, chen2020simple} based on the information maximization (InfoMax) principle has quickly gained popularity, leading significant improvements in learning unsupervised representations of natural images.
Specifically, it formulates an instance-wise classification problem; as the encoder learns to identify whether a pair of inputs is from the same sample or not, the mutual information between the representation and the data is maximized.  

In this work, we present a novel perspective for unsupervised representation learning: the encoder should attain maximum nonlinear expressivity on the data in order for downstream predictors to take full advantage of the encoder's nonlinear power. 
For a rectified activation (ReLU) network which is piece-wise linear, the nonlinear expressivity of the network is defined in terms of the number of distinct linear regions it defines on the input domain \citep{pascanu2013number, montufar2014number, raghu2017expressive}, where each linear region is associated with an \textit{activation pattern} of the encoder's hidden units. 
Based on this observation, \textit{neural activation coding} (NAC) maximizes the mutual information between the activation code and the data over a noisy communication channel.
We show that learning of a noise-robust activation code for communication increases the number of distinct linear regions of the network (\Cref{fig:linear_regions}) and therefore maximizes its nonlinear expressivity. 
Moreover, NAC learns \textit{both} continuous and discrete representations of data which we respectively evaluate on linear classification and nearest neighbor retrieval on natural image datasets.
Finally, we show that NAC pretraining improves the training of deep generative models by enhancing the encoder expressivity.  

Our main contributions are summarized as follows:
\begin{itemize}
	\item We propose \textit{neural activation coding} (NAC) as a novel approach for unsupervised representation learning. In contrast to contrastive learning approaches that are based on InfoMax principle, NAC maximizes the nonlinear expressivity of the encoder by formulating a communication problem over a noisy channel using the \textit{activation code} of the encoder.
	\item NAC is able to learn \textit{both} continuous and discrete representations of data, which we respectively evaluate on (i) linear classification on CIFAR-10 and ImageNet-1K and (ii) nearest neighbor search on CIFAR-10 and FLICKR-25K. We show NAC attains comparable or better performance to recent competitive methods, including SimCLR \citep{chen2020simple} and DistillHash \citep{yang2019distillhash}. 
	\item By maximizing the nonlinear expressivity of encoder, we demonstrate that NAC pretraining significantly benefits the training of variational autoencoders. 
	\item NAC does not require $\ell_2$-normalization for learning good representations, questioning the prevalent belief \citep{wu2018unsupervised, wang2020understanding} that $\ell_2$-normalization plays a key role in unsupervised representation learning. 
\end{itemize}

\section{Related Works}
\textbf{Nonlinear complexity of deep neural networks}. 
\citet{pascanu2013number, montufar2014number, raghu2017expressive, serra2018bounding, arora2018understanding, hanin2019complexity} have studied nonlinear expressivity of deep neural networks (DNNs). In particular, a network with rectified activation ($\text{ReLU}(x) = \max(0, x)$) is a piece-wise linear function and divides the input space distinct locally linear regions. 
Accordingly, the nonlinear expressivity of the network is represented by the number of distinct linear regions it defines on the input domain. 
In contrast to the previous works that have either sought theoretical bounds on the the number of linear regions in DNNs or empirically analyzed the nonlinear expressiveness of DNNs when being trained on a supervised task, we propose a way to explicitly maximize the nonlinear expressivity of the encoder for representation learning. 

\textbf{Self-supervised representation learning}. A prevalent approach for representation learning is to formulate \textit{pretext} tasks with pseudo labels generated from the unlabeled data \citep{doersch2015unsupervised, noroozi2016unsupervised, gidaris2018unsupervised}. 
Among others, contrastive learning \citep{wu2018unsupervised, oord2018representation, tian2019contrastive, he2020momentum, chen2020simple, chen2020improved} methods have recently led the state-of-the-art advances in linear classification and transfer learning on natural image datasets. 
Specifically, the approach of \citet{wu2018unsupervised, he2020momentum, chen2020simple} generates two different views of an example (e.g. random crops of an image); the views of the same example are treated as \textit{positive samples} while the views generated from distinct examples are treated as \textit{negative samples}. The encoder is trained to solve the instance discrimination problem of classifying the positive pair from the negatives. 
This maximizes the cosine similarity of positive pairs in the representation space, while pushing the representations of different examples away from each other.

Although contrastive representation learning has achieved significant advances on natural image data,  it is not yet well understood exactly why it has been so successful in learning representations for downstream predictors. One explanation is the information maximization (InfoMax) principle \citep{oord2018representation, bachman2019learning}, which states that the representation should contain maximum information about the data. Notably, contrastive learning optimizes a lower-bound to the mutual information (MI) between the representation and the data \citep{poole2019variational}. 
However, \citet{tschannen2020mutual} argue the success of contrastive approach cannot be attributed to the InfoMax principle alone but strongly rely on the properties of MI estimators and architectural choices. 
In this work, we present a new approach for representation learning that maximizes the nonlinear expressivity of the encoder for downstream predictors. 

\textbf{Unsupervised deep hashing}. 
Deep hashing aims to learn binary representations (i.e., \textit{hash codes}) of data using deep neural networks \citep{salakhutdinov2009semantic, krizhevsky2011using, lin2016learning, hu2017learning, yang2018semantic, yang2019distillhash}. 
The binary nature of the code admits efficient computation of nearest neighbor search algorithms for large-scale data, with minimal memory footprint. 
\citet{lin2016learning} treat images and their rotated ones as similar pairs and learn rotation invariant hash mapping. 
\citet{hu2017learning} is similar to NAC in that it maximizes the mutual information between the hash code and the data. However, they ignore higher-order interactions between the code bits in order to derive an approximation to the MI. On the other hand, NAC lower-bounds the MI using variational inference and subsampling, and further promotes noise-robustness of the code by introducing a noisy communication channel. 
More recently, \citet{yang2018semantic, yang2019distillhash} exploit similarity of deep features to construct pseudo labels for hash code learning and achieve the state-of-the-art performance on natural image datasets. 


\section{Approach}

\subsection{Activation Code in ReLU Networks}
Consider a deep neural network (DNN) with ReLU activation $\text{ReLU}(x) = \max(0, x)$:
\begin{align}
\vct{a}^{(l)} &= \W^{(l)} \h^{(l-1)} + \vct{b}^{(l)}, \label{eq:relu_network}\\ 
\h^{(l)} &= \text{ReLU}(\vct{a}^{(l)}), \quad l = 1, 2, \dots, L
\end{align}
where ReLU is applied element-wise and $L$ is the number of layers. We set $\h^{(0)} = \x$.
The DNN is a piece-wise linear function that segments the input space into a set of distinct locally linear regions (\Cref{fig:linear_regions}) \citep{pascanu2013number, montufar2014number, raghu2017expressive}. 

For layer $l$, we define the \textit{activation code} as the binary string
\begin{align}
\vct{c}^{(l)} = \sgn(\vct{a}^{(l)}) \in \{-1, 1\}^D, \label{eq:activation_code}
\end{align}
where $D$ is the number of hidden units in a layer. The activation code represents the \textit{activation pattern} of the network that uniquely identifies a linear region. Hereafter, we focus on the last layer activation code $\vct{c}^{(L)}$ and will drop the superscript when there is no ambiguity. 

The DNN maps each training data point $\x_i \in \{\x_1, ..., \x_N\}$ to an activation codeword $\vct{c}_i$ (\Crefrange{eq:relu_network}{eq:activation_code}).
The distance between two codewords $\vct{c}_i, \vct{c}_j \in \{-1, 1\}^D$ is measured using the \textit{Hamming distance}: 
\begin{align}
d_H(\vct{c}_i, \vct{c}_j) = \frac{D - \vct{c}_i \cdot \vct{c}_j}{2},
\end{align}
which counts the number of different bits between the two codewords. $\vct{c}_i \cdot \vct{c}_j$ denotes the dot product of the two codewords. The distance is the minimum number of distinct linear regions that one has to traverse along a path from $\x_i$ to $\x_j$. Therefore, the average distance or separability between the codewords serves as a measure of the effective number of linear regions on the data \citep{raghu2017expressive}. 

While we limit our attention to ReLU in this work, similar analyses may apply to a broad class of other activation functions. 
For example, Leaky ReLU and MaxOut are already piece-wise linear functions. Exponential linear units (ELUs) and Gaussian error linear units (GELUs) can be seen as smooth approximations to ReLU. Sigmoid and hyperbolic tangent (tanh) activations also have distinct modes of operation; they behave like a linear function near the origin but gradually saturate to constant functions further away from the origin. Hence, they can be considered as smooth approximations to piece-wise linear functions too. 

\subsection{Neural Activation Coding}
For a ReLU encoder, the distribution of that activation codewords $\{\vct{c}_1, ..., \vct{c}_N\}$ has significant implications for downstream applications as the nonlinear expressivity of the encoder is determined as the number of distinct activation codewords \citep{pascanu2013number, montufar2014number, raghu2017expressive},
For example, if a set of data are mapped to the same codeword on the DNN, it means that they lie in the same linear region of the encoder. Therefore, downstream linear models won't be able to express any nonlinear relationships between these examples. This suggests that a good encoder network should attain high nonlinear expressivity by mapping the data to as many unique activation codewords as possible.
This is the key motivation behind \textit{Neural Activation Coding} (NAC) which proposes to maximize the nonlinear expressivity of the encoder for representation learning.

NAC maximizes the mutual information (MI) between the \textit{activation code} and the data over a noisy communication channel $\vct{X} \rightarrow \vct{C} \rightarrow \widetilde{\vct{C}}$.
Suppose the message $\x_i \in \{\x_1, ..., \x_N\}$ is selected uniformly at random from the dataset. The sender first encodes the message $\x_i$ into the activation codeword $\vct{c}_i$ (\Crefrange{eq:relu_network}{eq:activation_code}) and transmits it through the noisy channel. 
The receiver tries to reconstruct the message from the noisy code $\widetilde{\vct{c}}_i$. 
In order for the receiver to correctly decode the message with high probability, the codewords $\{\vct{c}_1, ..., \vct{c}_N\}$ should be easily separable; in other words maximally distant from each other \citep{mackay2003information}. 
The amount of information that the receiver gains from the communication is quantified as the mutual information between the noisy activation code and the data: $I(\X, \widetilde{\vct{C}})$. 
Thus maximizing $I(\X, \widetilde{\vct{C}})$ leads to noise-robust codewords that are maximally distant from each other, which translates to maximum nonlinear expressivity of the the encoder. 

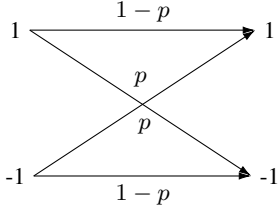
\begin{figure}[t]
	\centering
	\begin{tikzpicture}
	\small
	\node (upperright) {1};
	\node (upperleft) [left=3cm of upperright] {1}; 
	\node (lowerleft) [below=1.5cm of upperleft] {-1}; 
	\node (lowerright) [below=1.5cm of upperright] {-1};
	
	\draw [-{Latex[width=1mm]}] (upperleft.east) -- node [above]{$1-p$} (upperright.west);
	\draw [-{Latex[width=1mm]}] (upperleft.east) -- node [above, yshift=0.1cm]{$p$} (lowerright.west);
	\draw [-{Latex[width=1mm]}] (lowerleft.east) -- node [below, yshift=-0.1cm]{$p$} (upperright.west);
	\draw [-{Latex[width=1mm]}] (lowerleft.east) -- node [below]{$1-p$} (lowerright.west);
	\end{tikzpicture}
	\caption{A symmetric noise channel used in NAC. Each bit of code is independently flipped with probability $p$.}
	\label{fig:symmetric_channel}
\end{figure}

\textbf{Symmetric noise channel}. 
We consider a symmetric noise channel where the bits of $\vct{c}$ are randomly flipped with probability $p$ (\Cref{fig:symmetric_channel}) to create a noisy code $\tilde{\vct{c}}$. 
The total number of flipped bits is given by the Hamming distance $d_H(\tilde{\vct{c}}, \vct{c}) = (D - \tilde{\vct{c}} \cdot \vct{c}) / 2$.
The conditional probability of transmitted message $\tilde{\vct{c}}$ given a codeword $\vct{c}_i$ is
\begin{align}
&P(\tilde{\vct{c}}|\vct{c}_i) 
= p^{d_H(\tilde{\vct{c}}, \vct{c}_i)} (1-p)^{D - d_H(\tilde{\vct{c}}, \vct{c}_i)} \\
&= p^{(D - \tilde{\vct{c}} \cdot \vct{c}_i) / 2} (1-p)^{(D + \tilde{\vct{c}} \cdot \vct{c}_i) / 2} \\
&= \exp((\tilde{\vct{c}} \cdot \vct{c}_i) \frac{1}{2} \log\frac{1-p}{p} + \frac{D}{2} \log p(1 - p)).
\label{eq:symmetric_conditional}
\end{align}
Accordingly, the marginal distribution of the message is
\begin{align}
&P_{\theta}(\tilde{\vct{c}})
= \sum_{j, \vct{c}} P_{\text{data}}(\x_j) P_{\theta}(\vct{c} | \x_j) P(\tilde{\vct{c}} | \vct{c}) \label{eq:symmetric_marginal} \\ 
&= \frac{1}{N} \sum_{j=1}^N \exp((\tilde{\vct{c}} \cdot \vct{c}_j) \frac{1}{2} \log\frac{1-p}{p} + \frac{D}{2} \log p (1-p)), \nonumber
\end{align}
where $\theta$ denotes the parameters of the encoder and the data distribution is assumed to be uniform ($P_{\text{data}}(\x_j) = 1/N$) over the training examples.
The activation code is a deterministic function of the input i.e., $P_{\theta}(\vct{c}_j|\x_j) = 1$. 

The mutual information between the message and data is
\begin{align}
&I(\X, \widetilde{\vct{C}}) \label{eq:MI}
= \mathbb{E}_{\x, \vct{\tilde{c}}}\Bigg[ \log \frac{P_{\theta}(\tilde{\vct{c}} | \vct{c}_i)}{P_{\theta}(\tilde{\vct{c}})} \Bigg] \\
&= \frac{1}{N} \sum_{i=1}^N  \mathbb{E}_{\vct{\tilde{c}} | \vct{c}_i} \Bigg[ \log \frac{\exp((\tilde{\vct{c}} \cdot \vct{c}_i) \frac{1}{2} \log\frac{1 - p}{p})}{\frac{1}{N} \sum_{j} \exp((\tilde{\vct{c}} \cdot \vct{c}_j) \frac{1}{2} \log \frac{1 - p}{p})} \Bigg] \nonumber,
\end{align} 
which follows from \Cref{eq:symmetric_conditional,,eq:symmetric_marginal}. Note that $\log \frac{1 - p}{p} > 0$ for $p < 0.5$ (i.e., the channel has non-zero capacity).
From the denominator, we see maximizing $I(\X, \widetilde{\vct{C}})$ 
will minimize the similarity (i.e., maximize the Hamming distance) between the codewords and consequently  
maximize the number of distinct linear regions of the encoder. 

\subsection{Data Augmentation}
Data augmentations play a significant role in self-supervised learning of natural image representations \citep{chen2020simple, henaff2020data, wu2018unsupervised}. 
Common augmentation methods for images include horizontal flipping, random cropping, color jittering, etc.  
We incorporate data augmentations into NAC by modifying the communication channel as $\X \rightarrow \widetilde{\X} \rightarrow \vct{C} \rightarrow \widetilde{\vct{C}}$, where $\widetilde{\X}$ is the augmented version of $\X$ as a result of applying stochastic data augmentations. The mutual information is now
\begin{align}
I(\X, \widetilde{\vct{C}}) \label{eq:MI_augmented}
&= \mathbb{E}_{\x, \tilde{\vct{c}}} \Bigg[ \log \frac{P_{\theta}(\tilde{\vct{c}} | \x)}{P_{\theta}(\tilde{\vct{c}})} \Bigg].
\end{align}

Both the numerator and the denominator in \Cref{eq:MI_augmented} is no longer tractable since it requires marginalization over $\tilde{\x}$. We therefore construct lower-bounds for each term using (i) variational inference and (ii) subsampling, respectively as described below.

\textbf{Variational inference}. We lower-bound the numerator using an amortized variational distribution $Q_{\phi}(\tilde{\vct{c}}| \x)$:
\begin{align}
&\mathbb{E}_{\x, \tilde{\vct{c}}} [\log Q_{\phi}(\tilde{\vct{c}}| \x)] \label{eq:variational_bound} \\ 
&= \mathbb{E}_{\x, \tilde{\vct{c}}} [\log P_{\theta}(\tilde{\vct{c}} | \x)] - D_{KL}(P_{\theta}(\tilde{\vct{c}} | \x) \| Q_{\phi}(\tilde{\vct{c}}| \x))] \nonumber  \\
&\le \mathbb{E}_{\x, \tilde{\vct{c}}} [\log P_{\theta}(\tilde{\vct{c}} | \x)],
\end{align}
where the expectation is taken over $P_{\theta}(\x, \tilde{\vct{c}})$, and the inequality stems from the non-negativity of KL-divergence.
From the bound, see that maximizing \Cref{eq:variational_bound} in turn minimizes $D_{KL}(P_{\theta}(\tilde{\vct{c}} | \x) \| Q_{\phi}(\tilde{\vct{c}}| \x))]$, driving the variational distribution closer to the true conditional. 
We adopt a mean-field approach by setting $Q_{\phi}(\tilde{\vct{c}}| \x)$ as a product of Bernoulli distributions (mind that here $\tilde{c}_d \in \{-1, 1\}$):
\begin{align}
Q_{\phi}(\tilde{\vct{c}}| \x) 
= \prod_{d=1}^D q_d^{(1 + \tilde{c}_d) / 2} (1 - q_d)^{(1 - \tilde{c}_d) / 2}, \label{eq:variational_distribution}
\end{align}
where $D$ is the dimension of the code. 
We introduce an \textit{inference network} $\phi$ to output the logit $r_d = \log\frac{q_d}{1 - q_d}$ for each bit and then apply sigmoid function to obtain $q_d = \sigma(r_d)$. 
The expectation of \Cref{eq:variational_bound} is evaluated as
\begin{align}
&\mathbb{E}_{\x, \tilde{\vct{c}}} [\log Q_{\phi}(\tilde{\vct{c}}| \x)] \\
&= \mathbb{E}_{\x, \tilde{\vct{c}}} \Bigg[\frac{1}{2}   \sum_{d=1}^D \tilde{c}_d \log\frac{q_d}{1 - q_d} + \log q_d (1 - q_d)\Bigg] \\
&= \frac{1}{2} \mathbb{E}_{\x, \tilde{\vct{c}}} \Bigg[\tilde{\vct{c}} \cdot \vct{r}_{\phi}(\x) + \vct{1} \cdot \log \sigma(\vct{r}_{\phi}(\x)) (\vct{1} - \sigma(\vct{r}_{\phi}(\x))) \Bigg], \nonumber
\end{align}
where $\vct{r}_{\phi}(\x)$ is the $D$-dimensional logit vector, and $\vct{1}$ is the same-sized one vector. The sigmoid and the log functions are applied element-wise.  

\textbf{Subsampling}. On the other hand, the denominator of \Cref{eq:MI_augmented} can be lower-bounded using $2K$ subsamples $\vct{c}_{1}, \vct{c}_{2}, \dots, \vct{c}_{2K}$ \citep{poole2019variational, chen2020simple}. 
Specifically, we first sample $K$ examples $\x_1, \x_2, \dots, \x_K \sim P_{\text{data}}(\x)$ and draw two augmented versions per each example $\tilde{\x}_{2k-1}, \tilde{\x}_{2k} \sim P_{\text{aug}}(\tilde{\x}|\x_k)$. Finally, the encoder network maps the augmented samples $\tilde{\x}_1, \tilde{\x}_2, \dots, \tilde{\x}_{2K}$ to activation codewords $\vct{c}_{1}, \vct{c}_{2}, \dots, \vct{c}_{2K}$. The bound is constructed as
\begin{align}
&\mathbb{E}_{\tilde{\vct{c}}}\Bigg[\log \frac{1}{P_{\theta}(\tilde{\vct{c}})}\Bigg] \label{eq:subsample_bound} \\
&\ge \mathbb{E}_{\tilde{\vct{c}}, \vct{c}_{1}, \dots \vct{c}_{2K}}\Bigg[\log \frac{1}{\frac{1}{2K} \sum_{k=1}^{2K}  P(\tilde{\vct{c}} | \vct{c}_k)}\Bigg] \\
&= \mathbb{E}_{\tilde{\vct{c}}, \vct{c}_{1}, \dots \vct{c}_{2K}}\Bigg[\log \frac{1}{\frac{1}{2K} \sum_{k=1}^{2K}  \exp((\tilde{\vct{c}} \cdot \vct{c}_k) \frac{1}{2} \log \frac{1 - p}{p})} \Bigg]. \nonumber
\end{align}

\textbf{NAC objective}. Combining the two bounds above (\Cref{eq:variational_bound,eq:subsample_bound}), we arrive at our objective
\begin{align}
\mathcal{L}_{\text{NAC}}
&= \mathbb{E}_{\x, \tilde{\vct{c}}, \vct{c}_{1}, \dots, \vct{c}_{2K}} \Bigg[ \log \frac{Q_{\phi}(\tilde{\vct{c}} | \x)}{\frac{1}{2K} \sum_{k=1}^{2K}  P_{\theta}(\tilde{\vct{c}} | \vct{c}_k)} \Bigg], \label{eq:objective_augmented} 
\end{align}
which is a lower-bound to the mutual information $I(\X, \widetilde{\vct{C}})$.

\subsection{Optimization}
The objective of \Cref{eq:objective_augmented} involves discrete codewords and does not admit gradient-based optimization. To circumvent this issue, we adopt continuous relaxation \citep{cao2017hashnet} and replace the discrete code $\vct{c} = \sgn(\vct{a}) \in \{-1, 1\}^D$ with a soft approximation $\z = \text{tanh}(\vct{a}) \in [-1, 1]^D$ where $\vct{a}$ is the last preactivation of the encoder:
\begin{align}
&\widetilde{\mathcal{L}}_{\text{NAC}} \label{eq:optimization_objective} \\
&= \frac{1}{2} \mathbb{E}_{\x, \tilde{\vct{z}}} \Bigg[\tilde{\vct{z}} \cdot \vct{r}_{\phi}(\x) + \vct{1} \cdot \log \sigma(\vct{r}_{\phi}(\x)) (\vct{1} - \sigma(\vct{r}_{\phi}(\x))) \Bigg]  \nonumber \\
&\quad - \mathbb{E}_{\tilde{\vct{z}}, \vct{z}_{1}, \dots, \vct{z}_{2K}}[\log \frac{1}{2K} \sum_{k=1}^{2K}  \exp((\tilde{\vct{z}} \cdot \vct{z}_k) \frac{1}{2} \log \frac{1 - p}{p})]. \nonumber 
\end{align}
We apply stochastic gradient optimization by sampling a mini-batch of examples $\x_1, \x_2, \dots, \x_K \sim P_{\text{data}}(\x)$ at each iteration. The gradients with respect to
the parameters of the encoder $\theta$ and the inference network $\phi$ are computed using backpropagation on \Cref{eq:optimization_objective}. 

\subsection{Model Architecture}
\begin{figure}[t]
	\centering
	\includegraphics[width=0.5\textwidth]{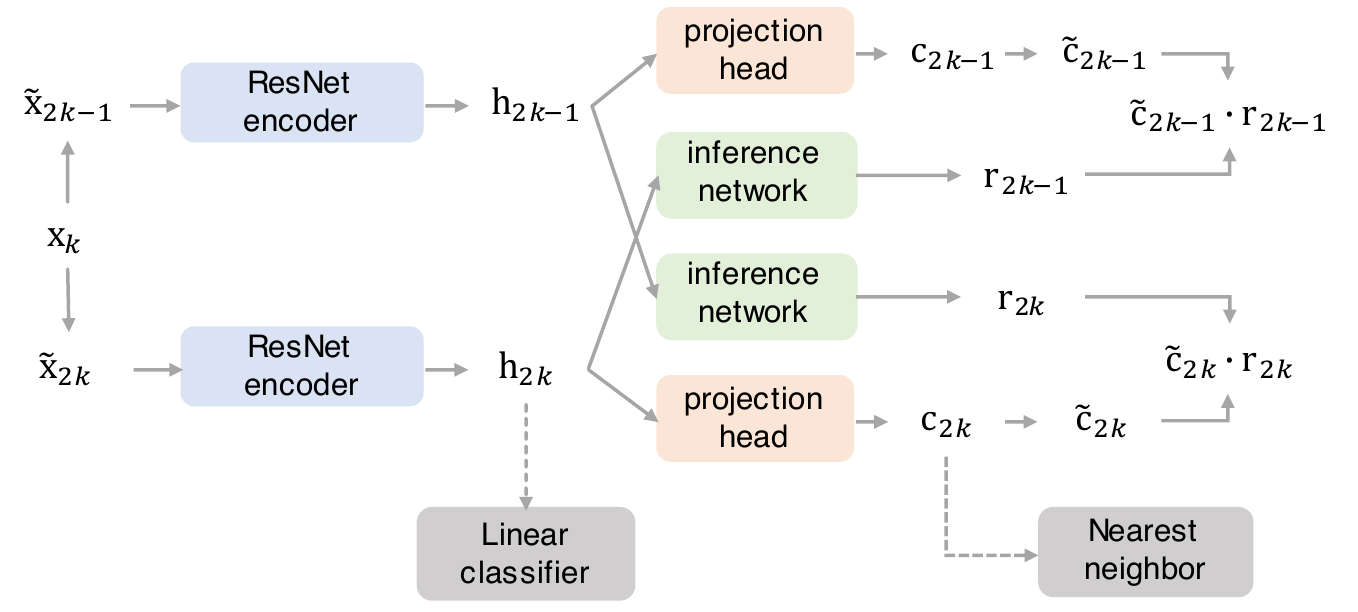}
	\caption{The NAC architecture. The inference network takes the encoder representation from one pathway and predict the logits for the other pathway. The negative samples are not depicted here.
		After training, the encoder representation and the activation code are respectively applied to linear classification and nearest neighbor search.}
	\label{fig:architecture}
\end{figure}	

\Cref{fig:architecture}  overviews the NAC architecture. 
The encoder takes the augmented data $\tilde{\x}$ as input and produces the representation $\h$. Following \citet{chen2020simple}, \citet{chen2020improved}, we attach a projection head at the end of the encoder. It is an MLP with one hidden layer that maps the encoder presentation $\vct{h}$ to the lower-dimensional feature $\vct{z}$. The activation code $\vct{c}$ is obtained by applying a sign function on $\vct{z}$.
Similarly, the inference network shares the same encoder backbone and predicts the logit vector $\vct{r}$ from the encoder representation $\vct{h}$. The logit $\vct{r}$ defines the variational distribution $Q_{\phi}(\tilde{\vct{c}}|\x)$ (\Cref{eq:variational_distribution}).
However,  directly using the encoder representation $\vct{h}$ from the same path allows the inference network to easily cheat.
Therefore, we build two independent pathways by sampling two augmented versions of data $\tilde{\x}_{2k-1}, \tilde{\x}_{2k} \sim P(\tilde{\x} | \x_k)$.
The inference network takes the encoder representation of one pathway (e.g., $\h_{2k-1}$) and outputs the logits for the other pathway (e.g. $\vct{r}_{2k}$) and vice versa (\Cref{fig:architecture}). 

For ImageNet experiments, we incorporate the momentum queue (MQ) \citep{he2020momentum} in order to reduce the memory overhead. The momentum queue maintains $M$ momentum features $\vct{v}_1, ..., \vct{v}_M$ from previous iterations. In addition, we introduce a momentum model $\widehat{\vct{r}}_{\phi}(\x)$ for the inference network as well. 
In our experiments, we find that the discrepancy between the norms of current model features and momentum features causes instability during the training of NAC-MQ models and add $\ell_2$-regularization term on the norm of the feature $\z$ in order to stabilize the training: 
\begin{align}
&\widetilde{\mathcal{L}}_{\text{NAC-MQ}} \label{eq:momentum_objective} \\
&= \frac{1}{2} \mathbb{E}_{\x, \tilde{\vct{z}}} \Bigg[\tilde{\vct{z}} \cdot \widehat{\vct{r}}_{\phi}(\x) + \vct{1} \cdot \log \sigma(\widehat{\vct{r}}_{\phi}(\x)) (\vct{1} - \sigma(\widehat{\vct{r}}_{\phi}(\x))) \Bigg]  \nonumber \\
&- \mathbb{E}_{\vct{z}, \tilde{\vct{z}}}\Bigg[\log \frac{1}{M} \sum_{m=1}^{M}  \exp((\tilde{\vct{z}} \cdot \vct{v}_m) \frac{1}{2} \log \frac{1 - p}{p}) + \lambda \|\z\|_2^2\Bigg], \nonumber 
\end{align}
where $\lambda$ controls the strength of $\ell_2$ regularization. We use $\lambda=0.1$ in our experiments. 

\subsection{Comparison to Contrastive Learning}
We compare NAC to the contrastive learning objective of SimCLR (\citet{chen2020simple}) and highlight a few distinguishing traits. 
The SimCLR objective for example $\x_i$ is
\begin{align}
&\mathcal{L}_{\text{SimCLR}}^{(i)} 
= \log \frac{\exp((\vct{u}_i \cdot \vct{u}'_i) / \tau)}{\sum_{k=1}^{2K} \mathbb{I}_{[k\neq i]} \exp((\vct{u}_i \cdot \vct{u}_k) / \tau)}, \label{eq:simclr}\\
&\text{ where } \vct{u}_k = \frac{\vct{z}_k}{\|\vct{z}_k\|} \text{ for } k=1, \dots, 2K. \label{eq:l2_normalization}
\end{align}
where $\vct{u}_i, \vct{u}_i'$ are the features of two augmented versions of the same image (i.e., a \textit{positive pair}) and $\vct{u}_1, \dots, \vct{u}_{2K}$ are the \textit{negative samples}. $\mathbb{I}$ is an indicator function. Notably, contrastive learning $\ell_2$-normalizes the features so that they lie on the unit hypersphere and use a temperature parameter $\tau$ to control the concentration of the distribution.

For comparison, we can rewrite the relaxed NAC objective in \Cref{eq:optimization_objective} as
\begin{align}
&\widetilde{\mathcal{L}}_{\text{NAC}}^{(i)}  \label{eq:nac_i}
= \log \frac{\exp((\tilde{\vct{z}}_i \cdot \vct{r}_{\phi}(\x_i)) / 2)}{\sum_{k=1}^{2K}  \exp((\tilde{\vct{z}}_i \cdot \vct{z}_k) \frac{1}{2} \log \frac{1 - p}{p})}.
\end{align}
We see that both methods take similar forms that incentivize minimizing the feature similarity between the negatives and maximizing the similarity for the positive pair.
However, SimCLR uses the sample feature of the same example as the positive, while NAC lets the inference network to predict the distribution $P_{\theta}(\tilde{\vct{z}} | \x)$ to guide the feature.

Moreover, NAC does not apply $\ell_2$-normalization to the features (\Cref{eq:l2_normalization}) but still learns good representations of data, even though the role of $\ell_2$-normalization has been considered crucial in self-supervised representation learning \citep{wu2018unsupervised, chen2020simple, wang2020understanding}.
We hypothesize that leaving out explicit normalization may be beneficial since it allows the model to represent uncertainty of its predictions in the norm of its features.
To see this, we rewrite the NAC objective (\Cref{eq:nac_i}) using the $\ell_2$-normalized features as
\begin{align}
\widetilde{\mathcal{L}}_{\text{NAC}}^{(i)}  &=\log \frac{\exp((\tilde{\vct{u}}_i \cdot \vct{r}_{\phi}(\x_i)) / 2)^{\|\tilde{\vct{z}}_i\|}}{\sum_{k=1}^{2K}  \exp((\tilde{\vct{u}}_i \cdot \vct{z}_k) \frac{1}{2} \log \frac{1 - p}{p})^{\|\tilde{\vct{z}}_i\|}} \label{eq:nac_softmax} \\
&\text{ where } \tilde{\vct{u}}_k = \frac{\tilde{\vct{z}}_k}{\|\tilde{\vct{z}}_k\|} \text{ for } k=1, \dots, 2K.
\end{align}
\Cref{eq:nac_softmax} can be interpreted as a softmax distribution where the norm of feature $\|\tilde{\vct{z}}_i\|$ dynamically controls the concentration of the distribution.
When the encoder is confident, it outputs large $\|\tilde{\vct{z}}_i\|$ to make the distribution sharp. 
Otherwise, it assigns small $\|\tilde{\vct{z}}_i\|$ to smooth the distribution reflecting its uncertainty.  
This is not possible for SimCLR that enforces the features to be on the unit hypersphere; thus, its performance is sensitive to tuning of the temperature parameter $\tau$ \citep{wu2018unsupervised, chen2020simple}. 

Finally, contrastive representation learning \citep{oord2018representation, poole2019variational, chen2020simple} maximizes the MI between the representation and the data $I(\X, \Z)$. In contrast, the goal of NAC is in maximizing the nonlinear expressivity of the encoder for downstream predictors. This is achieved by maximizing the MI between the activation code and the data $I(\X, \widetilde{\vct{C}})$ over a noisy communication channel. 
The noisy channel promotes noise-robust codewords which leads to high nonlinear encoder expressivity.

\section{Experiments}
We assess the quality of continuous and discrete representations learned by NAC on two downstream tasks respectively: (i) linear classification on CIFAR-10 and ImageNet-1K. (ii) nearest neighbor search using the activation hash code on CIFAR-10 and FLICKR-25. In addition, we explore whether deep generative models can benefit from enhanced encoder expressivity from NAC pretraining. We show that NAC attains better or comparable performance to recent methods including SimCLR \citep{chen2020simple} and DistillHash \citep{yang2019distillhash} on the downstream tasks and provides significant improvement for the training of variational autoencoders (VAEs) \citep{kingma14}. 


\subsection{Experimental Details}
Following previous works \citep{chen2020simple, he2020momentum, chen2020improved}, we use ResNet architecture with ReLU activation as our encoders. 
The projection head is an MLP with one hidden layer with ReLU activation. The feature/code dimension is set to 128. The inference network has the identical structure to the projection head.   
For optimization, we use LARS optimizer \citep{you2017large} with linear warmup for the first 10 epochs followed by cosine learning rate decay.  
We apply the same set of data augmentations including horizontal flipping, random cropping and resizing, color distortions and Gaussian blur used in 
\citet{chen2020simple, chen2020improved}.
We set weight decay to $10^{-6}$. For multi-GPU training, we adopt batch shuffling \citep{he2020momentum} to prevent the information leak in batch normalization layers. 

\textbf{CIFAR-10}.
We use a batch size of 1000 and train the encoder for 1000 epochs. Following \citet{chen2020simple}, we exclude Gaussian blur in CIFAR-10 experiments. 
The learning rate is set to 3.0 with momentum 0.9. 
For the models with momentum queue, we set the size of the queue to $50000$ and the moving average decay to $0.99$. 

\textbf{ImageNet}. 
We use a batch size of 512 and train the encoder for 200 epochs. 
The learning rate is set to 1.7 following the square root scaling rule ($0.075 \times \sqrt{\text{batch size}}$) \citep{chen2020simple} with momentum 0.9. 
For the models with momentum queue, we set size of the queue to $65536$ and the moving average decay to $0.999$, following \citep{he2020momentum}. 

\subsection{Linear Image Classification}
\begin{table}[t]
	\centering
	\caption{Linear evaluation accuracy (top-1) on CIFAR-10 dataset using ResNet-50 encoders. Trained for 1000 epochs.}
	\vspace{0.1in}
	\begin{threeparttable}
		\begin{tabular}{l c}
			\toprule
			Model	& Accuracy (\%) \\
			\midrule
			\textit{Contrastive Learning Methods:} & \\
			InsDis \citep{wu2018unsupervised} & 80.8 \tnote{*} \\
			SimCLR \citep{chen2020simple} & 92.8 \tnote{\textdagger} \\
			MoCo-v2 \citep{chen2020improved} &	91.6 \tnote{\textdagger} \\
			\midrule
			NAC & \textbf{93.9} \\
			NAC + Momentum Queue & 93.8 \\
			\bottomrule
		\end{tabular} 
		\begin{tablenotes}
			\footnotesize
			\item[*] Obtained using ResNet-18 encoder and kNN classifier 
			\item[\textdagger] Re-implemented for multi-GPU training.
		\end{tablenotes}
		\label{table:cifar10_accuracy}
	\end{threeparttable}
\end{table}

\begin{table}[t]
	\centering
	\caption{Linear evaluation accuracy (top-1) on ImageNet 1K dataset using ResNet-50 encoders. Trained for 200 epochs.}
	\vspace{0.1in}
	\begin{tabular}{l c}
		\toprule
		Model	& Accuracy (\%) \\
		\midrule
		\textit{Contrastive Learning Methods:} & \\
		InsDis \citep{wu2018unsupervised} & 54.0 \\
		CMC \citep{tian2019contrastive} & 60.0 \\
		LocalAgg \citep{zhuang2019local} & 60.2 \\
		Moco \citep{he2020momentum} & 60.6 \\
		SimCLR  \citep{chen2020simple} & 66.6\\
		Moco-v2 \citep{chen2020improved} & \textbf{67.5} \\
		\midrule
		NAC + Momentum Queue & 65.0 \\
		\bottomrule
	\end{tabular} 
	\label{table:imagenet_accuracy}
\end{table}

For downstream classification, the projection head is detached from the encoder, and the encoder representation is fed into a linear classifier. The encoder network is kept fixed and only the linear model is learned using the supervision. We measure the top-1 classification accuracy of the classifiers on the test set. It is a popular evaluation procedure for assessing the quality of continuous deep representations \citep{henaff2020data, he2020momentum, wu2018unsupervised}. 

The linear classifier is trained using Nesterov optimizer with momentum 0.9 for 100 epochs where the learning rate is searched among $\{0.01, 0.1, 1.0, 10.0\}$. 
We do not apply any regularization on the classifier.
For CIFAR-10 experiments, we re-implement SimCLR \citep{chen2020simple} and MoCo-v2 \citep{chen2020improved} for multi-GPU training and report the corresponding results for fair comparison.

\Cref{table:cifar10_accuracy} summarizes the downstream linear classification results on CIFAR-10. The NAC outperforms the state-of-the-art baselines by over 1\%p. We find that the MoCo-v2 attains slightly worse results than the SimCLR on CIFAR-10, while the NAC + Momentum Queue shows comparable performance to the vanilla NAC. 

\Cref{table:imagenet_accuracy} shows the linear classification results on ImageNet. NAC falls slightly behind the state-of-the-art baselines in ImageNet 1K
but there is room for additional improvements as we have not run extensive hyperparameter search due to computational constraints. 

\subsection{Nearest Neighbor Search Using Deep Hash Codes}
The goal of deep hashing is to utilize deep neural networks to learn binary vector code $\vct{c}$ (i.e., hash codes) of high-dimensional data $\x$ for efficient nearest neighbor retrieval \citep{krizhevsky12, erin2015deep, do2016learning}. 
Hamming distance is used for ranking and the binary nature of the code admits efficient nearest neighbor search for large-scale datasets with minimal memory footprint. 
The retrieved image is considered relevant if it belongs to the same class as the query image. 
The hash code table is populated using the training images and the test images are used as queries. 
The performance is measured using mean Average Precision (mAP) \citep{luo2020survey}, which computes the average area under the precision-recall curve. For NAC, the activation code of the projection head is used as the hash code. We compare NAC against recent deep hashing methods \citep{lin2016learning, yang2018semantic, yang2019distillhash} .

The hash code performance on CIFAR-10 is summarized in \Cref{table:cifar10_hashing}. 
For reference, we also evaluate the performance of contrastive methods of SimCLR \citep{chen2020improved} and MoCo-v2 \citep{chen2020improved} by discretizing the models' output using a sign function. Following \citet{yang2019distillhash}, we use a VGG16 encoder and train the models on 10\% of the dataset for fair comparison. We train the encoder from scratch, while the deep hashing baselines finetune a pretrained VGG16 encoder. 
NAC outperforms all the baselines by significant margins as it learns maximally \textit{separable} codewords by promoting noise-robustness. 
Interestingly, the contrastive learning methods of SimCLR and Moco-v2 surpass the deep hashing baselines, even though they are not explicitly designed for learning hash codes. This may be attributed to the use of strong data augmentations that these models incorporate during training.  

\Cref{table:flickr25k_hashing} shows the results on FLICKR-25K. Following \citet{yang2019distillhash}, we start from a VGG16 encoder pretrained on ImageNet-1K classification and finetune the model using NAC. Against the deep hashing baselines, NAC attains the highest mAP. However, as ImageNet pretraining already provides a strong baseline for FLICKR-25K, we find that the performance margins are not as significant as in the CIFAR-10 results. 



\begin{table}[t]
	\centering
	\caption{Retrieval performance of unsupervised hash code (128 bit) on CIFAR-10. The baselines results for deep hashing methods are excerpted from \citep{yang2019distillhash}. The models are trained on 10\% of the data using VGG16 encoders.} 
	\vspace{0.1in}
	\begin{tabular}{l c}
		\toprule
		Model	& mAP (\%) \\
		\midrule
		
		\textit{Deep hashing methods} &\\
		DeepBit \citep{lin2016learning}  & 25.3 \\
		SSDH \citep{yang2018semantic} &  26.0 \\
		DistillHash \citep{yang2019distillhash}	& 29.0 \\
		\midrule
		\textit{Contrastive learning methods} &\\
		MoCo-v2 \citep{chen2020improved} & 32.3 \\
		SimCLR \citep{chen2020simple} & 34.2 \\
		\midrule
		NAC & \textbf{40.5} \\
		\bottomrule
	\end{tabular} 
	\label{table:cifar10_hashing}
\end{table}

\begin{table}[t]
	\centering
	\caption{Retrieval performance of unsupervised hash code (128 bit) on FLICKR-25K. The baselines results for deep hashing methods are excerpted from \citep{yang2019distillhash}. The models are trained on 5000 images using VGG16 encoders pretrained on ImageNet-1K.} 
	\vspace{0.1in}
	\begin{tabular}{l c}
		\toprule
		Model	& mAP (\%) \\
		\midrule
		DeepBit \citep{lin2016learning}  & 59.3 \\
		SSDH \citep{yang2018semantic} &  66.2 \\
		DistillHash \citep{yang2019distillhash}	& 70.0 \\
		\midrule
		NAC & \textbf{70.8} \\
		\bottomrule
	\end{tabular} 
	\label{table:flickr25k_hashing}
\end{table}


\subsection{Encoder Pretraining for Deep Generative Models}
\begin{table}[t]
	\centering
	\caption{Comparison of VAE performance using random initialization and unsupervised pretraining on CIFAR-10. The loglikelihoods are estimated with importance sampling. Trained for 100 epochs.} 
	\vspace{0.1in}
	\begin{tabular}{l r r}
	\toprule
	Encoder & Loglikelihood & KL divergence  \\
	\midrule
	Random init. & -3202 & 33.0 \\
	SimCLR + finetune & -3174 & 38.9 \\
	MoCo-v2 + finetune & -3103 & 32.2 \\
	NAC	+ finetune  & \textbf{-2865} & \textbf{71.8} \\
	\bottomrule
\end{tabular} 
	\label{table:vae}
\end{table}

The deep generative models of variational autoencoders (VAEs) \citep{kingma14} take the encoder-decoder architecture where the decoder defines the generative distribution given a latent variable, and the encoder predicts the posterior distribution of the latent variable. 
However, VAEs suffer from suboptimality of the encoder \citep{cremer18, marino18, kim18}, which leads to biased learning signals. 
This problem is exacerbated by several factors, namely: (i) the encoder is randomly initialized, leading to the \textit{cold start} problem. (ii) the encoder is never trained to the optimality. This may be mitigated with additional optimization steps, but they come with significant computational overheads. (iii) To make the problem worse, the learning target for the encoder constantly changes as the decoder is jointly updated with the encoder.

Given that even linear models can learn complex functions when combined with pretrained encoders, 
we hypothesize that NAC pretraining for the encoder can benefit VAEs by maximizing the encoder's nonlinear expressivity. 
Specifically, we pretrain the encoder using NAC and only randomly initialize the linear output layer to predict the posterior distribution of the latent variable.
This allows us to apply higher learning rates on the linear output layer to speed up the training of the encoder and consequently improve the quality of amortized inference. 
We also finetune the encoder by backpropagating through the linear output layer. 

\Cref{table:vae} summarizes the results on CIFAR-10 using ResNet-18 architecture. We compare the NAC pretraining against random initialization and pretraining using self-supervised methods. Using the NAC pretraining achieves significantly higher loglikelihood and KL divergence compared to the baselines. 
This suggests that maximizing the encoder expressivity using NAC facilitates more active use of latent variable and significantly enhances the training of VAEs.  

\subsection{The Effect of Noisy Communication Channel}
\begin{figure}[t]
	\centering
	\includegraphics[width=0.45\textwidth]{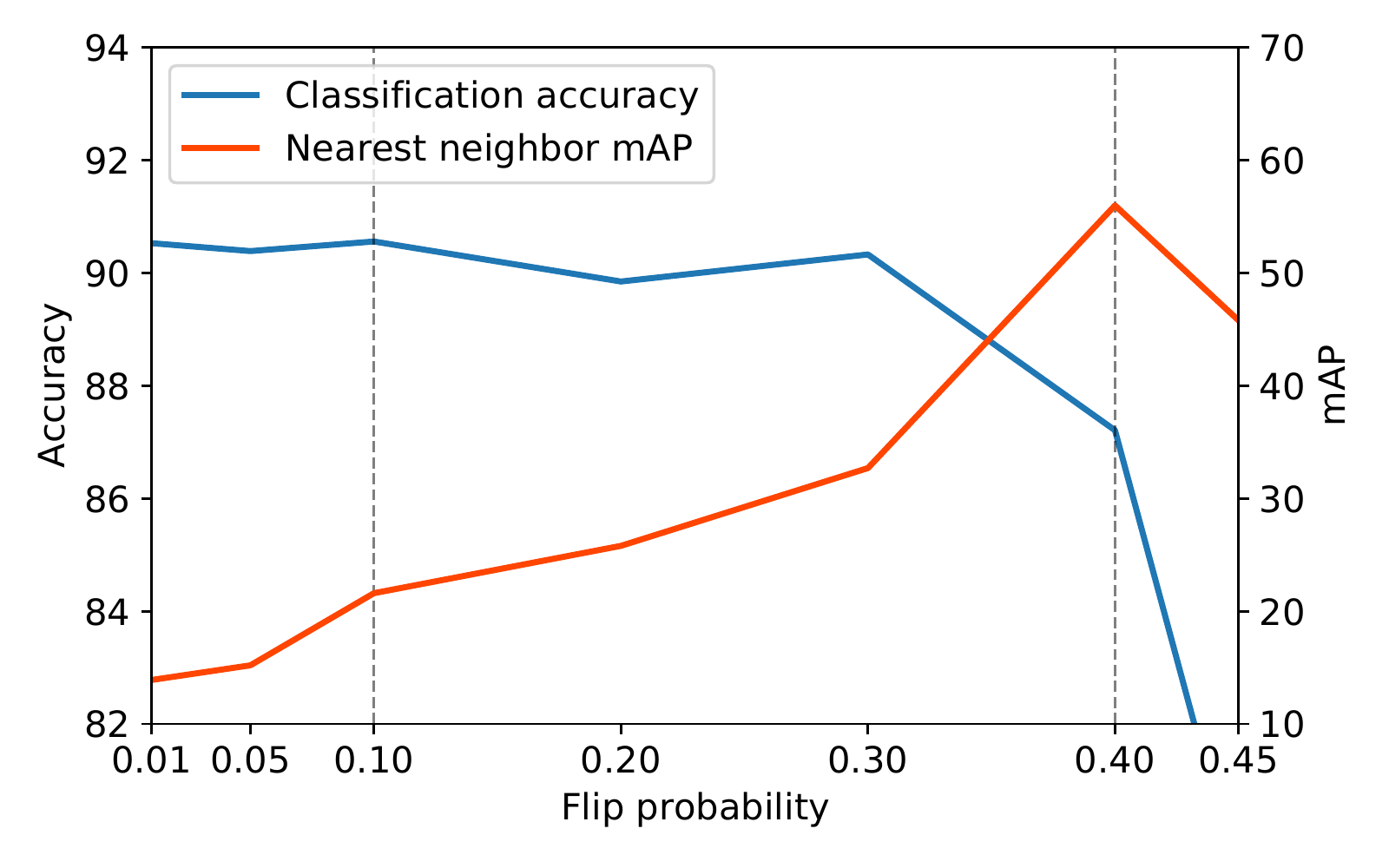}
	\vspace{-0.1in}
	\caption{CIFAR-10 downstream performance using ResNet-18 encoders with different flip probabilities. The dashed lines denote the values of $p$ that attains the best accuracy on respective tasks.}
	\label{fig:flip_probability}
\end{figure}
  
The noisy communication channel in NAC controls the overall difficulty of task with the bit flip probability $p$ of the channel.
\Cref{fig:flip_probability} plots the effect of the flip probability on the downstream classification and nearest neighbor search on CIFAR-10 . 
Interestingly, while the smaller noise probability of $p=0.1$ is favorable for linear classification, the nearest neighbor performance peaks at the higher noise level of $p=0.4$, suggesting discrete hash code benefits more from improved noise-robustness. On the other hand, the continuous representation suffers as the task becomes too difficult, agreeing with the findings of \citet{chen2020simple}.
Note that in the limit $p\rightarrow 0.5$, the transmitted code becomes completely random, making the task impossible to solve. 

\section{Conclusion}
We proposed Neural Activation Coding (NAC) for unsupervised representation learning. NAC maximizes the nonlinear expressivity of the encoder by formulating a communication problem over a noisy communication channel using the activation code of the encoder. Through empirical evaluations, we demonstrated that NAC can improve the performance of downstream applications as well as enhance the training of deep generative models.
As future work, it is worthwhile to explore the use of NAC on other data domains such as natural language.

\section*{Acknowledgements}
This work is supported by ONR N00014-17-1-2131, ONR N00014-15-1-2209, NIH 1U01MH115727-01, NSF CCF-1740833, DARPA SD2 FA8750-18-C-0130, Amazon and Simons Foundation.
Sangho Lee and Gunhee Kim are supported by Institute of Information \& communications Technology Planning \& Evaluation (IITP) grant funded by the Korea government (MSIT) (No.2017-0-01772, Video Turing Test, No.2019-0-01082, SW StarLab).

We thank Christian A. Naesseth for helpful discussion.

\bibliography{references}
\bibliographystyle{icml2021}

\end{document}


\twocolumn[
\icmltitle{[Appendix] \\Unsupervised Representation Learning via Neural Activation Coding}



\icmlsetsymbol{equal}{*}

\begin{icmlauthorlist}
\icmlauthor{Yookoon Park}{columbia}
\icmlauthor{Sangho Lee}{snu}
\icmlauthor{Gunhee Kim}{snu}
\icmlauthor{David M. Blei}{columbia}
\end{icmlauthorlist}

\icmlaffiliation{columbia}{Computer Science Department, Columbia University, New York, USA}
\icmlaffiliation{snu}{Department of Computer Science and Engineering,  Seoul National University, Seoul, South Korea}

\icmlcorrespondingauthor{David M. Blei}{david.blei@columbia.edu}

\icmlkeywords{Machine Learning, ICML}

\vskip 0.3in
]



\printAffiliationsAndNotice{}  

\section{Experimental Details for VAE Pretraining}
We use diagonal Gaussians for both the variational posterior and the generative distribution of VAEs.  
For the encoder, we attach a linear output layer on ResNet-18 to predict the mean and the variances of a Gaussian distribution. The decoder takes a similar architecture with transposed convolution layers.  
We jointly train the encoder and the decoder on CIFAR-10 for 100 epochs with a batch size of 128.
We use the Adam optimizer~\citep{kingma15} with $\beta_1=0.9$, $\beta_2=0.999$, $\epsilon=1e-8$ and no weight decay.
The global learning rate is set to 5e-4. When finetuning, we apply a smaller learning rate of 5e-5 to the pretrained ResNet encoder, while keeping the learning rate high for the linear output layer. 


\section{MI and Average Hamming Distance} 
We argue that maximizing the mutual information (MI) $I(\X, \widetilde{\vct{C}})$ over a noisy communication channel learns the codewords that have high Hamming distance to each other. We here show a relationship between the mutual information and the average Hamming distance between the codewords. Specifically, the mutual information \textit{lower-bounds} the average Hamming distance. 
Recall that the Hamming distance between two codewords $\vct{c}_i, \vct{c}_j \in \{-1, 1\}^D$ is
\begin{align}
d_H(\vct{c}_i, \vct{c}_j) = D - \frac{\vct{c}_i \cdot \vct{c}_j}{2}.
\end{align}
The average Hamming distance is defined as 
\begin{align}
\overline{d_H} = \frac{1}{N(N-1)}\sum_{j \neq i} d_H(\vct{c}_i, \vct{c}_j).
\end{align}
Let $\tilde{\vct{c}}$ be the noisy message transmitted through a binary symmetric channel with flip probability $p$. Then
\begin{align}
\mathbb{E}_{\tilde{\vct{c}}|\vct{c}_i}[\tilde{\vct{c}} \cdot \vct{c}_j] = (1 - 2p)(\vct{c}_i \cdot \vct{c}_j).
\end{align}
Finally,
\begin{alignat}{2}
&I(\X, \widetilde{\vct{C}}) && \\
&= \frac{1}{N} \sum_{i=1}^N &&\mathbb{E}_{\tilde{\vct{c}}|\vct{c}_i}\Bigg[ \log \frac{\exp((\tilde{\vct{c}} \cdot \vct{c}_i) \frac{1}{2} \log\frac{1 - p}{p})}{\frac{1}{N} \sum_{j=1}^N \exp((\tilde{\vct{c}} \cdot \vct{c}_j) \frac{1}{2} \log \frac{1 - p}{p})} \Bigg]  \nonumber \\
&= \frac{1}{N} \sum_{i=1}^N \Bigg(&&\mathbb{E}_{\tilde{\vct{c}}|\vct{c}_i}[(\tilde{\vct{c}} \cdot \vct{c}_i) \frac{1}{2} \log\frac{1 - p}{p}] \\
& &&-\mathbb{E}_{\tilde{\vct{c}}|\vct{c}_i}[\log \frac{1}{N} \sum_{j=1}^N \exp((\tilde{\vct{c}} \cdot \vct{c}_j) \frac{1}{2} \log \frac{1 - p}{p})]\Bigg) \nonumber \\
&= \frac{1}{N} \sum_{i=1}^N \Bigg(&&\frac{D(1 - 2p) }{2} \log\frac{1 - p}{p} \\
& &&- \mathbb{E}_{\tilde{\vct{c}}|\vct{c}_i}[\log \frac{1}{N} \sum_{j=1}^N \exp((\tilde{\vct{c}} \cdot \vct{c}_j) \frac{1}{2} \log \frac{1 - p}{p})]\Bigg) \nonumber \\
&\le \frac{1}{N} \sum_{i=1}^N \Bigg(&&\text{Const} - \mathbb{E}_{\tilde{\vct{c}}|\vct{c}_i}[\frac{1}{N} \sum_{j=1}^N (\tilde{\vct{c}} \cdot \vct{c}_j) \frac{1}{2} \log \frac{1 - p}{p}]\Bigg) \tag{Jensen's inequality} \\
&= \frac{1}{N} \sum_{i=1}^N \Bigg(&&\text{Const} + \sum_{j \neq i} (1 - 2p)(\vct{c}_i \cdot \vct{c}_j)) \frac{1}{2} \log \frac{1 - p}{p}\Bigg) \nonumber \\
&= \frac{N-1}{N} (1&&-2p)  \log(\frac{1 - p}{p}) \overline{d_H} + \text{Const}.
\end{alignat}
\begin{align}
&\le \frac{1}{N} \sum_{i=1}^N \Bigg(\text{Const} - \mathbb{E}_{\tilde{\vct{c}}|\vct{c}_i}[\frac{1}{N} \sum_{j=1}^N (\tilde{\vct{c}} \cdot \vct{c}_j) \frac{1}{2} \log \frac{1 - p}{p}]\Bigg) \tag{Jensen's inequality} \\
&= \frac{1}{N} \sum_{i=1}^N \Bigg(\text{Const} + \sum_{j \neq i} (1 - 2p)(\vct{c}_i \cdot \vct{c}_j)) \frac{1}{2} \log \frac{1 - p}{p}\Bigg) \nonumber \\
&= \frac{N-1}{N} (1-2p)  \log(\frac{1 - p}{p}) \overline{d_H} + \text{Const}.
\end{align}
Therefore, maximizing the mutual information objective increases the average Hamming distance between the codewords.

\bibliography{references}
\bibliographystyle{icml2021}